\documentclass{article}

\PassOptionsToPackage{numbers,compress}{natbib}
\usepackage[preprint]{neurips_2026}

\usepackage[utf8]{inputenc} 
\usepackage[T1]{fontenc}    
\usepackage{hyperref}       
\usepackage{url}            
\usepackage{booktabs}       
\usepackage{amsfonts}       
\usepackage{nicefrac}       
\usepackage{microtype}      
\usepackage{xcolor}         
\usepackage{comment}
\usepackage{algorithm}
\usepackage{algorithmic}

\usepackage{float} 

\definecolor{codegreen}{rgb}{0,0.6,0}

\usepackage{amsmath}
\usepackage{amssymb}
\usepackage{mathtools}
\usepackage{arydshln}
\usepackage{float}
\usepackage{subcaption} 
\usepackage{amsthm}
\usepackage{pifont} 
\usepackage[dvipsnames]{xcolor}

\usepackage[capitalize,noabbrev]{cleveref}

\theoremstyle{plain}

\theoremstyle{definition}

\theoremstyle{remark}

\usepackage[textsize=tiny]{todonotes}
\usepackage{enumitem}
\usepackage{multirow}
\usepackage{makecell}

\newcommand{\passup}[1]{{\tiny\textcolor{green!50!black}{\textbf{+#1}}}}
\newcommand{\passdown}[1]{{\tiny\textcolor{red!70!black}{\textbf{-#1}}}}
\newcommand{\sysname}{\textsc{CORVUS}}
\newcommand{\decrease}[1]{{\tiny\textcolor{ForestGreen}{\textbf{$\downarrow$#1}}}}
\newcommand{\increase}[1]{{\tiny\textcolor{red}{\textbf{$\uparrow$#1}}}}

\newcommand{\nodiff}
{\phantom{\decrease{00.0\%}}}

\title{\sysname: \underline{C}ontext \underline{O}ptimization and \underline{R}eduction \underline{V}ia \underline{U}nderlying \underline{S}ynchronization for LLM Coding Agents}

\author{%
  \textbf{Mingwei Zheng}$^{1,*,\dagger}$\quad
  \textbf{David O'Brien}$^{2,*}$\quad
  \textbf{Siwei Cui}$^{2}$\quad
  \textbf{Pardis Pashakhanloo}$^{2}$, \\
  \textbf{Rajdeep Mukherjee}$^{2}$\quad
  \textbf{Myeongsoo Kim}$^{2}$\quad
  \textbf{Sachit Kuhar}$^{2}$ \\[0.4em]
  $^{1}$Department of Computer Science, Purdue University \\
  $^{2}$AWS AI Labs \\
  \texttt{zheng618@purdue.edu} \\
  \texttt{\{dmobrien, siweicui, ppardis, mukherr, mysoo, skuhar\}@amazon.com}
}

\begin{document}
\maketitle
\begingroup

\renewcommand{\thefootnote}{}

\footnotetext{$^{*}$Equal contribution. \quad
$^{\dagger}$Work done during an internship at Amazon.}

\endgroup



\begin{abstract}
LLM coding agents operate by constructing trajectories that accumulate reasoning, tool calls, and results to enable multi-step decision-making. However, the conventional append-only trajectory architecture found in practice tightly couples file-read actions with their observations, capturing snapshots that become permanently fixed in the chronological history. As files change through agent edits or concurrent human modifications, these snapshots become stale, causing reasoning errors and causing agents to redundantly re-read files, with each re-read appending yet another copy to the trajectory. To mitigate this, we propose \sysname{}, a novel trajectory architecture that decouples file-read actions from their observations by maintaining a synchronized registry of relevant files and injecting only their \emph{current} contents at each reasoning cycle. This structural change produces significantly lighter-weight trajectories that remain synchronized with the actual codebase state by construction, eliminating redundant file copies and stale snapshots that bloat conventional trajectories. We evaluated \sysname{} on \textsc{SWE-PolyBench\_Verified} and \textsc{SWE-Bench Pro} across four LLMs, achieving 9--50\% reduction in average input tokens per task, 15--32\% shorter final prompts, and up to 37\% fewer reasoning cycles while maintaining comparable pass rates.
\end{abstract}

\section{Introduction}
\label{sec:intro}
Large language model (LLM) coding agents such as Cursor~\cite{cursor}, Claude Code~\cite{claudecode}, and Kiro~\cite{kiro} have demonstrated strong capabilities in end-to-end software engineering tasks.
These agents typically follow an iterative \emph{Thought–Action–Result} loop: the model reasons about the task, invokes tools to inspect or modify the repository, observes the results, and uses the accumulated interaction history to choose the next action. This history, or \emph{trajectory}, serves as the agent’s working memory. It records prior reasoning, tool calls, file reads, edits, command outputs, and error messages, enabling the agent to perform long-horizon, multi-step development workflows.

\textbf{Context Rot in Coding Agents.} Despite their effectiveness, conventional append-only trajectories are inefficient for long-horizon coding tasks.
Each reasoning cycle appends new messages, tool calls, and observations to the trajectory, causing the agent's working context to grow rapidly. 
This growth increases token cost and inference latency, and may also reduce reasoning reliability: prior work shows that long contexts can make models less effective at locating and using relevant information~\cite{ DBLP:conf/emnlp/DuTRRBGWSHP25,DBLP:journals/tacl/LiuLHPBPL24, DBLP:journals/corr/abs-2502-05167}; accuracy often degrades as prompts grow, a phenomenon known as context rot~\cite{contextrot}.
Thus, trajectory growth is both an efficiency bottleneck and a reliability concern for long-running coding agents.


\begin{figure}[t]
\includegraphics[width=\columnwidth]{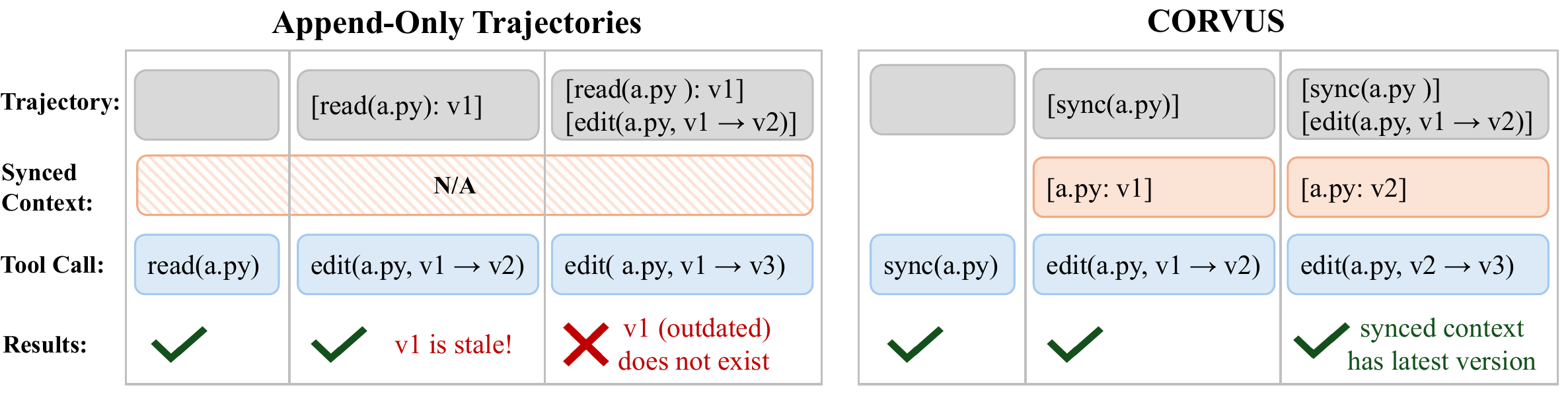}
\centering
\caption{Append-only trajectories accumulate stale snapshots, causing redundant reads and erroneous writes. \sysname{} maintains synchronized context, eliminating these inefficiencies by design.}
\label{fig:main-idea}
\end{figure}

\textbf{Motivation.} 
Existing context-management methods primarily reduce context after it has already accumulated~\cite{Strands-contextwindow, OpenHands,cursor-summarization,DBLP:journals/corr/abs-2508-08322}.
Although effective at shortening long trajectories, these methods are fundamentally \emph{reactive}: they operate on context that has already entered the trajectory, rather than preventing unnecessary or stale content from being introduced in the first place.
We target trajectory growth at its source by examining how context enters an agent’s history.
Our empirical analysis shows that a major source of inefficiency is how file reads are represented.
As shown on the left of \Cref{fig:main-idea}, conventional append-only trajectories tightly couple each file read action with its observation, storing a point-in-time snapshot of the file in the chronological history. Once recorded, this snapshot is never updated, even when later edits modify the underlying file. As a result, the trajectory gradually becomes desynchronized from the current codebase state and accumulates duplicate or outdated file contents, leading to two inefficiencies:

\begin{enumerate}[leftmargin=*, itemsep=0pt]
    \item \textbf{Duplicate file reads.} In our empirical analysis, 26\% of \textsc{Claude Sonnet 4} file-read operations re-read files that were already read in earlier cycles. Each re-read appends another full file snapshot to the message history, bloating the trajectory with duplicate content.
    
    \item \textbf{Stale-context errors.} Agents may edit based on outdated file snapshots preserved in the trajectory, e.g., replacing text that no longer exists, causing the edit to fail. Such failures require additional diagnosis and recovery cycles, further expanding the trajectory.
\end{enumerate}

\textbf{Our System.} We propose \sysname{}, an efficiency-oriented trajectory architecture that proactively synchronizes agent context with the repository state (\Cref{fig:main-idea}, right).
The key idea is that file contents are mutable and therefore should not be stored as fixed snapshots in append-only history.
\sysname{} implements this by replacing inline file reads with a \texttt{sync\_file} tool: when invoked, the tool registers a file in a \emph{synced file set} rather than appending its contents to the trajectory. Before each reasoning cycle, a \emph{context sync} phase refreshes all registered files from the repository and injects their current contents into the prompt. This design keeps at most one current version of each synced file in context, while ensuring that the agent reasons over the latest codebase state.

\textbf{Evaluation.} We extensively evaluate \sysname\ on \textsc{SWE-PolyBench\_Verified} and \textsc{SWE-Bench Pro} using four state-of-the-art LLMs spanning open-source and proprietary models. \sysname\ produces substantially lighter-weight trajectories: 9--50\% reduction in average input tokens per task, 15--32\% shorter final prompts, up to 86\% fewer duplicate file reads, and up to 37\% fewer reasoning cycles, all while maintaining comparable pass rates to baseline agents. These leaner trajectories also yield  up to 51\% reduced token costs and up to 40\% lower end-to-end latency. These results demonstrate that proactive codebase-trajectory synchronization is effective for enabling efficient agentic solutions for long-horizon software development tasks.

\textbf{Contributions.} In summary, this paper makes the following contributions:

\begin{itemize}[leftmargin=*, itemsep=0pt]
    \item We empirically identify a \emph{structural} inefficiency in LLM coding agents: the append-only trajectory architecture stores mutable file contents as immutable history entries, causing duplicate and stale file snapshots to accumulate.
    
    \item We propose \sysname{}, an efficiency-oriented trajectory architecture that keeps file contents synchronized with the repository instead of storing them as fixed history snapshots. \sysname{} maintains a synced file set and refreshes registered files before each reasoning cycle, ensuring that the prompt contains at most one current copy of each relevant file.
    
    \item We implement \sysname\ on \textsc{Strands Agents} and evaluate it on \textsc{SWE-PolyBench\_Verified} and \textsc{SWE-Bench Pro}. 
    Across four LLMs, \sysname{} reduces average input tokens by 9--50\%, shortens final prompts by 15--32\%, and cuts reasoning cycles by up to 37\%, while preserving comparable pass rates and remaining complementary to reactive context-management methods.
\end{itemize}

\section{Related Work}\label{sec:related_work}

\textbf{Long-Context Degradation.}
Growing evidence shows that longer trajectories harm not only cost but also model performance.
Liu et al.~\cite{DBLP:journals/tacl/LiuLHPBPL24} show that LLMs often underuse information in the middle of long prompts.
NoLiMa~\cite{DBLP:journals/corr/abs-2502-05167} further suggests that effective context length can be much shorter than nominal context length.
Du et al.~\cite{DBLP:conf/emnlp/DuTRRBGWSHP25} show that increasing input length can degrade performance even with perfect retrieval.
Hong et al.~\cite{contextrot} characterize this accuracy decay as context rot and quantify it across modern models.
Together, these findings motivate context management for long-horizon agents: even when trajectories fit within the nominal context window, unnecessary, duplicate, or stale context can make the model less effective at using the information it needs.


\textbf{Context Management for Long-Horizon Agents.}
Prior work mitigates trajectory growth in coding agents primarily through reactive trajectory management: the agent first accumulates reasoning steps, tool calls, and observations, and a separate mechanism later reduces the accumulated context~\cite{DBLP:journals/corr/abs-2508-08322,DBLP:journals/corr/abs-2603-05344}. Sliding-window methods retain recent turns and truncate older history once the prompt exceeds a context budget~\cite{Strands-contextwindow}. Observation masking and context editing selectively omit tool outputs or messages deemed less relevant~\cite{claudecode-contextediting, DBLP:journals/corr/abs-2508-21433}.
Another line of work uses additional LLMs to rewrite accumulated trajectories: summarization-based approaches periodically replace earlier history with compact natural-language summaries of the agent’s progress and state~\cite{OpenHands, cursor-summarization}.
AgentDiet identifies useless, redundant, or expired information and removes/replaces those parts~\cite{DBLP:journals/corr/abs-2509-23586}.
ACON~\cite{ACON} unifies observation and history compression by optimizing compression guidelines in natural language.
A complementary line of work reduces trajectory growth indirectly by steering execution: process reward models discourage inefficient reasoning paths~\cite{SWEPRM}, and multi-attempt refinement frameworks recover from failed runs ~\cite{SEAgent}.
In contrast, \sysname{} addresses trajectory growth \textbf{proactively}: rather than compressing a trajectory after redundant or outdated file content has accumulated, \sysname{} prevents duplicate and stale snapshots from being stored at all. \sysname{} is therefore complementary with reactive methods: reactive methods compress downstream while \sysname{} prevents a major source of growth upstream. We empirically confirm this complementarity in \Cref{subsec:compatibility}.

\section{Motivation}\label{sec:motiv}

\subsection{Problem Formulation}

Let $\mathcal{F} = \{f_1, f_2, \ldots, f_n\}$ denote the set of files in a codebase, where each file $f_i$ has content $c_i(t)$ at time $t$. An LLM coding agent operates through a sequential decision-making process over discrete time steps $t \in \{1, 2, \ldots, T\}$.

Traditional LLM coding agents found in practice operate through a standard agentic loop where each iteration generates a reasoning step $r_i$, executes an action $a_i$, and observes the result $o_i$. The agent maintains a trajectory $\mathcal{H}_t = \{(r_1, a_1, o_1), (r_2, a_2, o_2), \ldots, (r_t, a_t, o_t)\}$ of its past tool calls and results, where each triple $(r_i, a_i, o_i)$ is appended to the trajectory. For file-read actions $a_i = \texttt{read\_file}(f_j)$, the observation $o_i = c_j(t_i)$ captures the complete file content at the time of reading $t_i$, which becomes embedded in the trajectory.

This trajectory grows monotonically, and we refer to this as the \emph{append-only} trajectory architecture. Some systems apply \emph{reactive} transformations $\phi: \mathcal{H}_t \rightarrow \mathcal{H}'_t$ that compress or prune the trajectory after it exceeds some threshold~\cite{Strands-contextwindow,OpenHands, claudecode-contextediting}. However, these interventions are triggered only when context limits are approached and do not address the fundamental coupling between file-read actions and their point-in-time observations. In this work, we take a \emph{proactive} approach: rather than requiring explicit post-processing functions to clean up inefficiencies after they accumulate, we modify how trajectories are constructed so that efficiency is implicit in the architecture itself.

\subsection{Inefficiency Analysis}

Our empirical analysis of coding-agent trajectories reveals two key inefficiencies inherent to the append-only trajectory architecture:

\textbf{Observation 1: Duplicate File Reads.}
When a file $f_j$ is read multiple times, the trajectory contains multiple snapshots $\{c_j(t_{i_1}), c_j(t_{i_2}), \ldots, c_j(t_{i_k})\}$ for reading times $t_{i_1} < t_{i_2} < \ldots < t_{i_k}$. Append-only trajectories tightly couple file-read tool calls with their results, offering no mechanism to keep historical snapshots synchronized with the current codebase. This causes frequent re-reads: in our experiments on \textsc{SWE-PolyBench\_Verified} with Claude Sonnet 4, 26\% of file-read actions (834 out of 3,207) were duplicates. Re-reads typically occur when earlier content becomes buried in the growing prompt due to the ``lost in the middle'' phenomenon~\cite{DBLP:journals/tacl/LiuLHPBPL24}, when files change due to agent or external edits, or when agents verify that modifications were applied. Each re-read appends another full copy, compounding context growth and accelerating context rot~\cite{contextrot}.

\textbf{Observation 2: Incorrect Writes Due to Stale Context.}
If file $f_j$ is modified at time $t'$ after being read at time $t < t'$, the trajectory retains outdated content $c_j(t)$ while the current state is $c_j(t')$. Agentic actions are highly interdependent, and outdated file contents in the trajectory mislead downstream edits. Agents may reference code structures, variable names, or line numbers that no longer exist, causing write operations to fail. Recovery then requires re-reading files and incurs additional token overhead, as illustrated in \Cref{fig:main-idea}.

\section{\sysname}
\sysname{} addresses the inefficiencies identified in \Cref{sec:motiv} through a structural change to trajectory architecture. \Cref{alg:corvus} formalizes the \sysname{} agent loop, which augments the standard agentic workflow with a dedicated context synchronization phase.

\begin{algorithm}[H]
\caption{\sysname{} Agent Loop}
\label{alg:corvus}
\small
\setlength{\tabcolsep}{3pt}
\renewcommand{\arraystretch}{1.05}

\newcommand{\algind}{\hspace{1.2em}}
\newcommand{\algindd}{\hspace{2.4em}}

\begin{tabular}{r@{\quad}l@{\qquad} !{\color{gray!35}\vrule width 0.4pt} @{\qquad}r@{\quad}l}
1  & \textbf{Initialize:} $\mathcal{S}_0 \leftarrow \emptyset,\ \mathcal{H}_0 \leftarrow \emptyset$
   & 7  & \algindd $\mathcal{S}_t \leftarrow \mathcal{S}_{t-1} \cup \{f_j\}$ \\

2  & \textbf{for} $t = 1$ to $T$ \textbf{do}
   & 8  & \algindd $o_t \leftarrow \texttt{"sync:"}\ f_j$ \\

3  & \algind $\mathcal{C}_t \leftarrow \texttt{sync\_context}(\mathcal{S}_{t-1})$
   & 9  & \algind \textbf{else} \\

   & \algind \textcolor{gray}{\textit{// Context Sync}}
   & 10 & \algindd $o_t \leftarrow \texttt{execute\_action}(a_t)$ \\

4  & \algind $\mathcal{P}_t \leftarrow
        \mathcal{M}_{\text{sys}} \oplus
        \mathcal{T}_{\text{schema}} \oplus
        \mathcal{H}_{t-1} \oplus
        \mathcal{C}_t$
   & 11 & \algindd $\mathcal{S}_t \leftarrow \mathcal{S}_{t-1}$ \\

   & \algind \textcolor{gray}{\textit{// Compose Prompt}}
   & 12 & \algind \textbf{end if} \\

5  & \algind $(r_t, a_t) \leftarrow \texttt{llm\_invoke}(\mathcal{P}_t)$
   & 13 & \algind $\mathcal{H}_t \leftarrow \mathcal{H}_{t-1} \cup \{(r_t, a_t, o_t)\}$ \\

6  & \algind \textbf{if} $a_t = \texttt{sync\_file}(f_j)$ \textbf{then}
   & 14 & \textbf{end for}
\end{tabular}

\end{algorithm}

\begin{figure*}[t!]
\includegraphics[width=\linewidth]{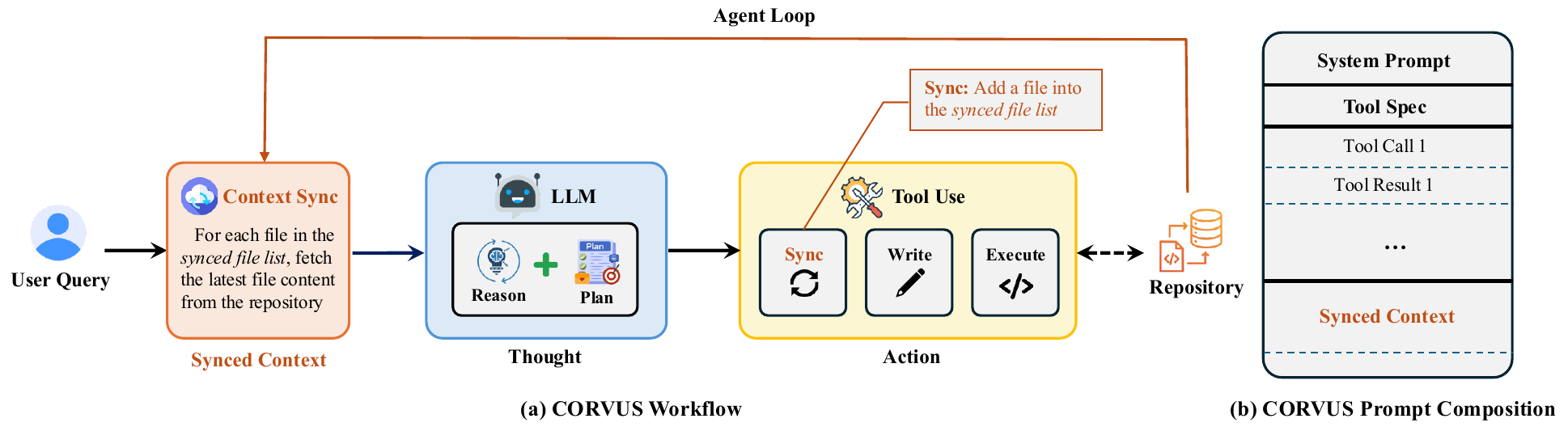}
\centering
\vspace{-5mm}
\caption{\sysname{} augments the standard \emph{Thought–Action–Result} loop with a \emph{Context Sync} phase (orange) that refreshes the agent's context by fetching the latest contents of all relevant files before each reasoning step. The resulting \emph{synced context} is appended to each prompt (orange), ensuring every reasoning cycle operates on an up-to-date repository snapshot.}
\label{fig:corvus}
\end{figure*}

\subsection{Synced File Set}

\sysname{} maintains a dynamic registry $\mathcal{S}_t \subseteq \mathcal{F}$ of relevant files (Line 1, \Cref{alg:corvus}). The set is updated through a new tool $\texttt{sync\_file}(f_j)$ (Lines 6-7) introduced in \sysname{}, where $\mathcal{S}_{t} = \mathcal{S}_{t-1} \cup \{f_j\}$ when $a_{t} = \texttt{sync\_file}(f_j)$.

The key innovation is that $\texttt{sync\_file}$ decouples file registration from content retrieval. When invoked, it produces a lightweight marker $m_j = \texttt{"sync:"}f_j$ as the observation $o_{t}$ (Line 8), while simultaneously updating the synced file set as a side effect. The trajectory records the triple $(r_{t}, a_{t}, o_{t})$ with only the compact marker (Line 13), avoiding the immediate inclusion of file contents.

\subsection{Context Synchronization}

At each reasoning step $t$, \sysname{} performs a context synchronization operation (Line 3) that constructs the synced context: $\mathcal{C}_t = \{(f_j, c_j(t)) : f_j \in \mathcal{S}_{t-1}\}$.

This ensures that the agent always reasons over the current file states $\{c_j(t)\}$ rather than historical snapshots. \Cref{fig:corvus}(a) illustrates this workflow: \sysname{} augments the standard \emph{Thought–Action–Result} loop with a \emph{Context Sync} phase at the beginning of each cycle.

\subsection{Prompt Composition}

The complete prompt $\mathcal{P}_t$ at time $t$ is structured as shown in Line 4 of \Cref{alg:corvus}: $\mathcal{P}_t = \mathcal{M}_{\text{sys}} \oplus \mathcal{T}_{\text{schema}} \oplus \mathcal{H}_{t-1} \oplus \mathcal{C}_t$, where $\mathcal{M}_{\text{sys}}$ is the system message, $\mathcal{T}_{\text{schema}}$ contains tool schemas, $\mathcal{H}_{t-1}$ is the message history, and $\oplus$ denotes concatenation. \Cref{fig:corvus}(b) shows how \sysname{} structures the prompt. By placing file contents in a dedicated \emph{synced context} block rather than scattered throughout the chronological message history, \sysname{} guarantees that (1) at most one version of each file appears in the prompt and (2) that version always reflects the current repository state.

\begin{figure*}[t]
\includegraphics[width=\linewidth]{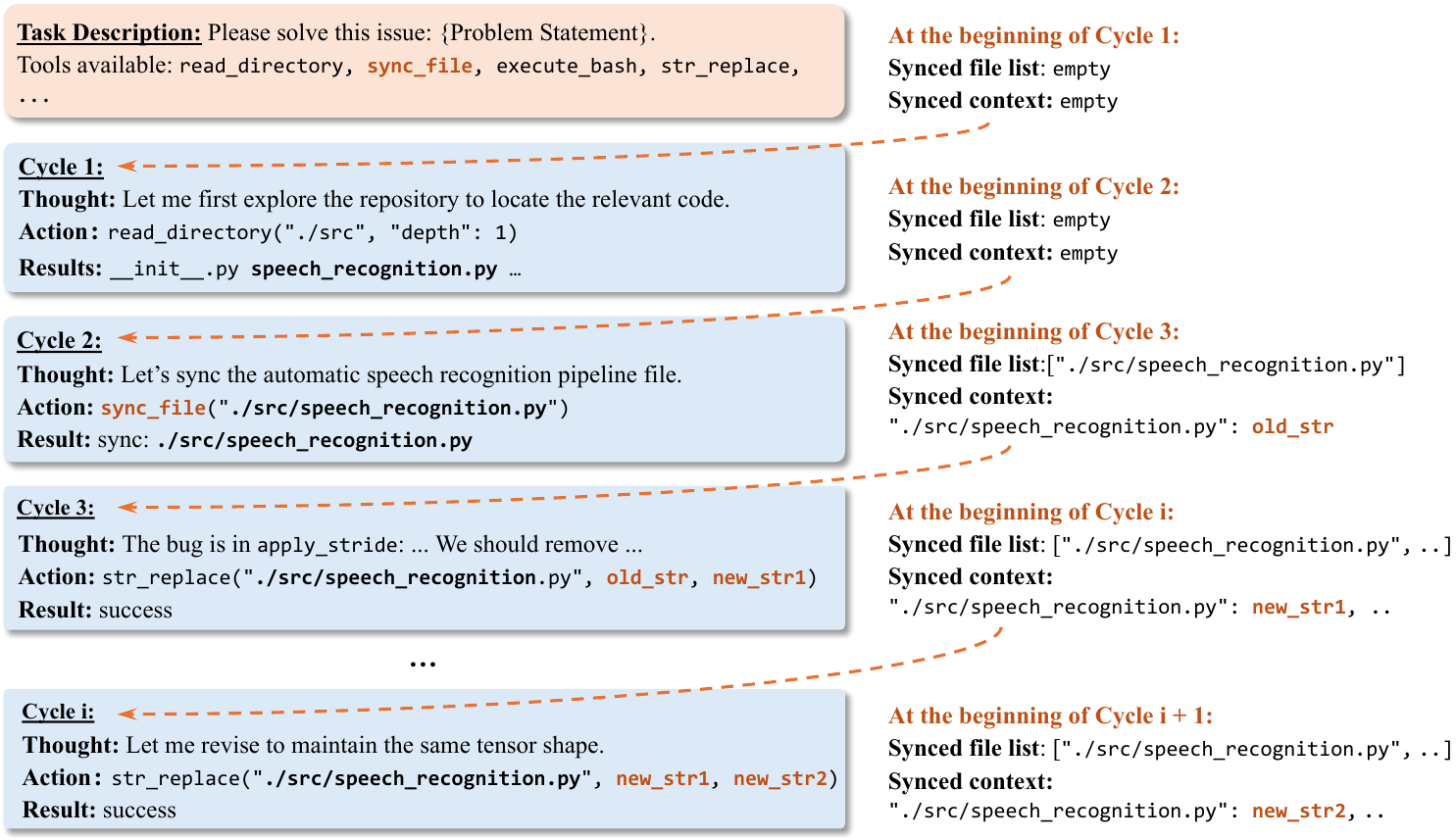}
\centering
\caption{Example: \sysname{} on \texttt{huggingface\_transformers-15843}.}
\label{fig:case}
\end{figure*}

\paragraph{Example.} \Cref{fig:case} shows \sysname{} on a real task.
The baseline agent reads \textit{speech\_recognition.py} in Cycle~2, modifies it in Cycle~3, then attempts another edit using the stale Cycle~2 snapshot, causing an edit failure. With \sysname{}, \texttt{sync\_file} registers the file in Cycle~2 with only a marker, while \emph{Context Sync} refreshes its contents before each subsequent cycle, enabling all edits to succeed.

\section{Implementation}\label{sec:impl}
We implement \sysname{} using the \textsc{Strands Agents} framework~\cite{Strands}, a library for building LLM agents with custom tools (5.8K GitHub stars at time of writing).
To maximize the generalizability and industry relevance of our evaluation, we survey the tools supported by 14 production and research coding agents (\Cref{tab:agents}) and derive a Minimum Viable Agent (MVA) from their shared tools: file read/write/edit, directory listing, grep search, and command execution. Our baseline implements these capabilities with \texttt{read\_file}, \texttt{fs\_write}, \texttt{list\_directory}, and \texttt{execute\_bash}.
\sysname{} changes only file reading: it replaces \texttt{read\_file} with \texttt{sync\_file} and adds a context-sync phase before each reasoning cycle, appending the refreshed synced context to the end of the prompt.

\begin{table}[!htb]
\centering
\small
\caption{Tools supported as of Jan. 23, 2026 by existing coding agents. Symbols denote support levels: \checkmark\ indicates full support, o denotes optional or partial support, and $\times$ indicates no support.}
\vspace{3mm}
\label{tab:agents}
\setlength{\tabcolsep}{4pt}
\renewcommand{\arraystretch}{1.05}
\newcommand{\rh}[1]{\rotatebox[origin=lB]{80}{\footnotesize\textsc{#1}~}}
\newcommand{\rhh}[2]{\rotatebox[origin=lB]{80}{\footnotesize\shortstack[l]{\textsc{#1}\\\textsc{#2}}~}}
\begin{tabular}{l|cccccccc|cccc|cc|cl}
\toprule
& \multicolumn{8}{c|}{\textit{Production: IDE-based}}
& \multicolumn{4}{c|}{\textit{Production: CLI-based}}
& \multicolumn{2}{c|}{\textit{Research}}
& \textit{MVA} & \\
\cmidrule(lr){2-9} \cmidrule(lr){10-13} \cmidrule(lr){14-15} \cmidrule(lr){16-16}
\textbf{Tool}
& \rh{Cursor}
& \rh{Kiro}
& \rh{Windsurf}
& \rh{Copilot}
& \rh{Cline}
& \rh{Roo Code}
& \rh{Continue}
& \rh{Zed}
& \rhh{Claude}{Code}
& \rhh{Codex}{CLI}
& \rh{Aider}
& \rh{Goose}
& \rhh{SWE}{Agent}
& \rhh{Open}{Hands}
& \rh{MVA} & \\
\midrule
Read File    & \checkmark & \checkmark & \checkmark & \checkmark & \checkmark & \checkmark & \checkmark & \checkmark & \checkmark & \checkmark & \checkmark & \checkmark & \checkmark & \checkmark & \checkmark & \\
Write File   & \checkmark & \checkmark & \checkmark & \checkmark & \checkmark & \checkmark & \checkmark & \checkmark & \checkmark & \checkmark & \checkmark & \checkmark & \checkmark & \checkmark & \checkmark & \\
Edit File    & \checkmark & \checkmark & \checkmark & \checkmark & \checkmark & \checkmark & \checkmark & \checkmark & \checkmark & \checkmark & \checkmark & \checkmark & \checkmark & \checkmark & \checkmark & \\
List Dir     & \checkmark & \checkmark & \checkmark & \checkmark & \checkmark & \checkmark & \checkmark & \checkmark & \checkmark & \checkmark & \checkmark & \checkmark & \checkmark & \checkmark & \checkmark & \\
Grep Search  & \checkmark & \checkmark & \checkmark & \checkmark & \checkmark & \checkmark & \checkmark & \checkmark & \checkmark & \checkmark & \checkmark & \checkmark & \checkmark & \checkmark & \checkmark & \\
AST Search   & \checkmark & $\times$   & \checkmark & \checkmark & \checkmark & \checkmark & \checkmark & o          & $\times$   & $\times$   & \checkmark & $\times$   & $\times$   & o          & $\times$   & \\
Emb. Search  & \checkmark & o          & \checkmark & o          & $\times$   & \checkmark & \checkmark & o          & $\times$   & \checkmark & $\times$   & $\times$   & $\times$   & $\times$   & $\times$   & \\
Exec. Cmd    & \checkmark & \checkmark & \checkmark & \checkmark & \checkmark & \checkmark & \checkmark & \checkmark & \checkmark & \checkmark & \checkmark & \checkmark & \checkmark & \checkmark & \checkmark & \\
Browser      & \checkmark & $\times$   & \checkmark & \checkmark & \checkmark & \checkmark & $\times$   & $\times$   & o          & o          & o          & $\times$   & $\times$   & \checkmark & $\times$   & \\
Fetch URL    & o          & \checkmark & o          & \checkmark & \checkmark & \checkmark & \checkmark & \checkmark & \checkmark & o          & \checkmark & o          & $\times$   & \checkmark & $\times$   & \\
MCP Tools    & \checkmark & \checkmark & \checkmark & \checkmark & \checkmark & \checkmark & \checkmark & \checkmark & \checkmark & \checkmark & $\times$   & \checkmark & $\times$   & \checkmark & $\times$   & \\
Diagnostics  & \checkmark & \checkmark & \checkmark & \checkmark & \checkmark & \checkmark & o          & \checkmark & \checkmark & \checkmark & \checkmark & \checkmark & \checkmark & $\times$   & $\times$   & \\
\bottomrule
\end{tabular}
\end{table}



\section{Evaluation}\label{sec:evaluation}


\subsection{Experimental Setup}\label{subsec:expr_setup}

\textbf{Benchmarks.}
We evaluate \sysname{} on two benchmarks that extend the popular \textsc{SWE-Bench}~\cite{swebench} in complementary directions. \textsc{SWE-PolyBench\_Verified}~\cite{polyverified} broadens language coverage beyond Python to Java, JavaScript, and TypeScript while adding feature addition and refactoring tasks; it contains 382 verified instances, each paired with ground-truth patches. \textsc{SWE-Bench Pro}~\cite{swebenchpro} targets enterprise-grade complexity with tasks requiring hours to days for professional engineers, averaging 107.4 lines across 4.1 files; due to evaluation cost, we sample instances with above-average code changes, yielding 191 valid cases after filtering invalid commits. This selection targets long-horizon tasks with extensive file interactions where context growth is most problematic, stress-testing \sysname{}'s efficiency claims. In total, our experiments cover 573 instances.

\textbf{Models.}
We evaluate on four LLMs: three proprietary models (\textsc{Claude Sonnet 3.7}~\cite{anthropic_claude_3_7_sonnet}, \textsc{Claude Sonnet 4}~\cite{anthropic_claude_4_sonnet}, \textsc{Claude Sonnet 4.5}~\cite{anthropic_claude_4.5_sonnet}) and one open-source model (\textsc{Qwen3-Coder-480B-A35B}~\cite{qwen3_coder}). This selection spans commercial and open-source models, assessing \sysname{}'s generality. We set temperature to 0.0 for all models.

\textbf{Baseline.}
To isolate the effect of synchronized context, we compare \sysname{} against the MVA agent without synchronization. The baseline embeds \texttt{read\_file} outputs directly into the trajectory, while \sysname{} uses \texttt{sync\_file} to register files for refresh before each reasoning cycle. All other tools, prompts, and configurations are identical.

\subsection{Trajectory Efficiency}
\textbf{Setup and Metrics.} We evaluate whether \sysname{} produces lighter-weight trajectories using four metrics:
(1)~\emph{Duplicate file reads}, the number of \texttt{read\_file} or \texttt{sync\_file} calls targeting a file already accessed in a previous cycle. This metric quantifies redundant file access patterns that bloat trajectories.
(2)~\emph{Reasoning cycles}, the number of iterations the agent requires to complete a task. Cycle count reflects the effort the model expends during problem-solving, including recovery from errors such as failed edits caused by stale context.
(3)~\emph{Final request length}, the token count of the last prompt sent to the LLM before task completion. Because the trajectory grows monotonically across cycles, the final request represents the largest prompt the agent constructs.
(4)~\emph{Accumulated token usage}, the total input and output tokens across all reasoning cycles, along with the corresponding monetary cost.
All results are averaged over all instances on each benchmark under identical configurations.

\begin{figure*}[!htb]
    \centering
      
    \begin{minipage}[t]{0.325\textwidth}
        \centering
        \includegraphics[width=\linewidth]{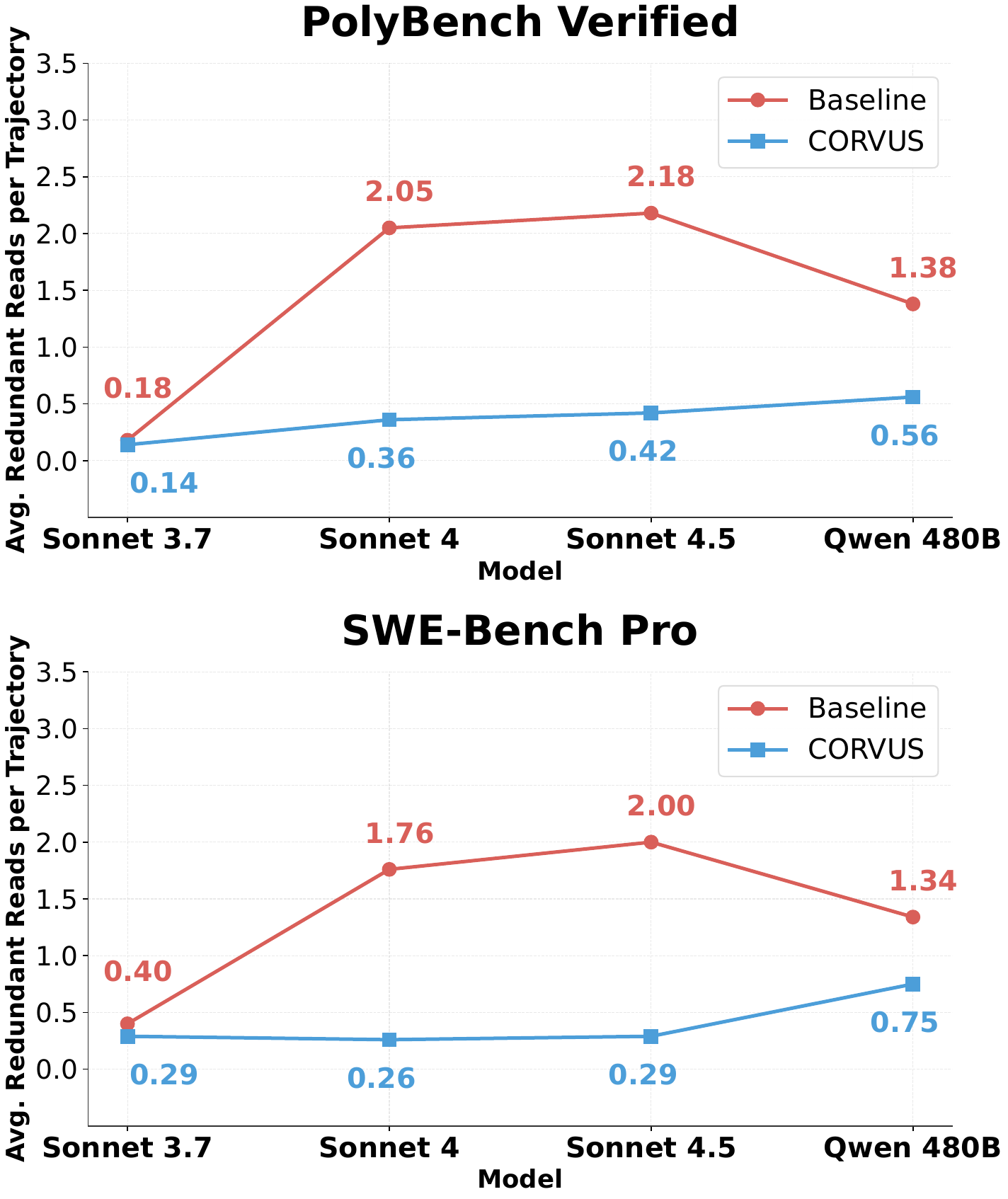}
        \vspace{-5mm}
        \caption{Duplicate file reads.}
        \vspace{-2mm}
        \label{fig:reason}
    \end{minipage}
    \hfill 
    \begin{minipage}[t]{0.327\textwidth}
        \centering
        \includegraphics[width=\linewidth]{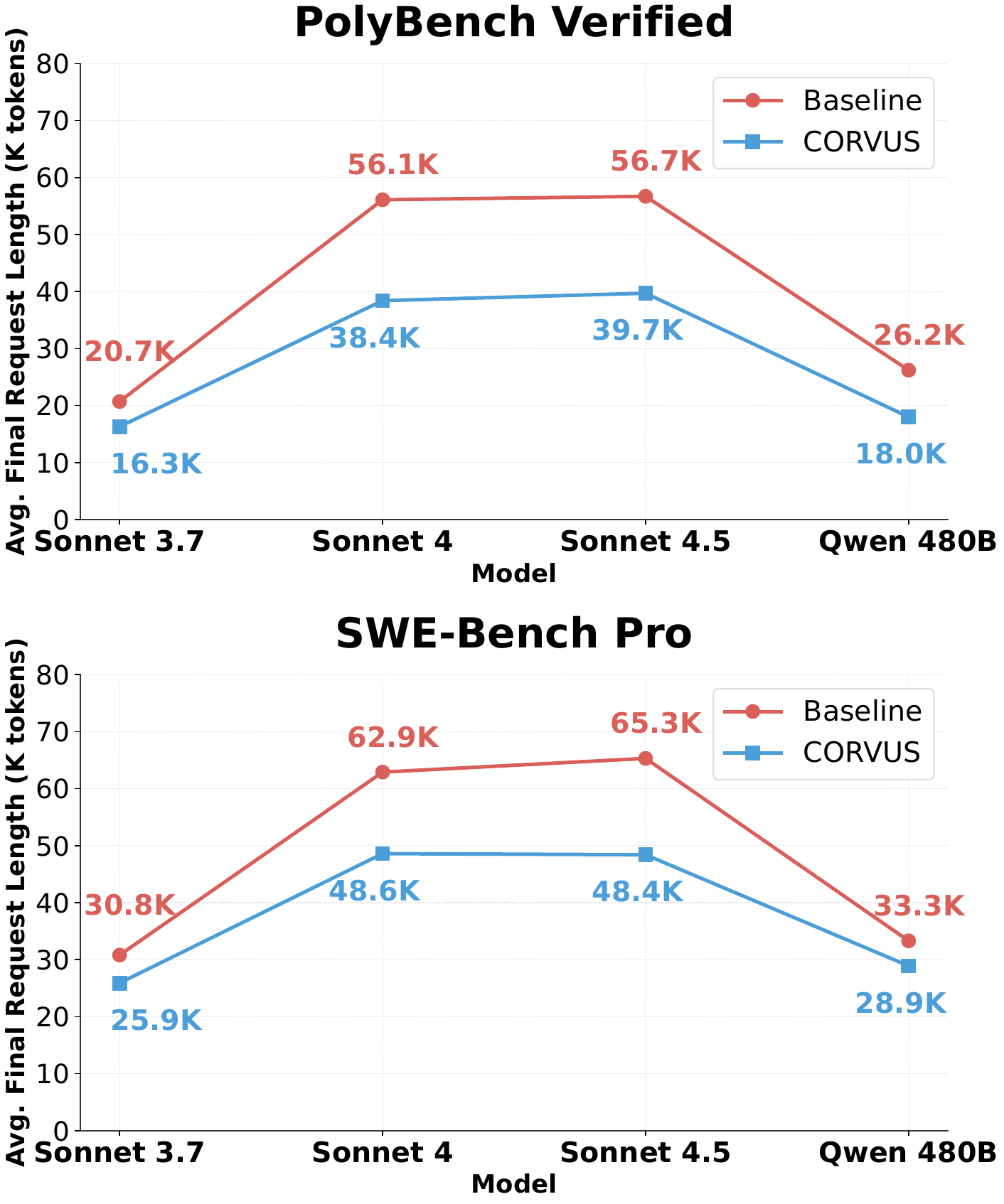}
        \vspace{-5mm}
        \caption{Final-request length.}
        \vspace{-2mm}
        \label{fig:finalrequest}
    \end{minipage}
    \hfill
    \begin{minipage}[t]{0.325\textwidth}
        \centering
        \includegraphics[width=\linewidth]{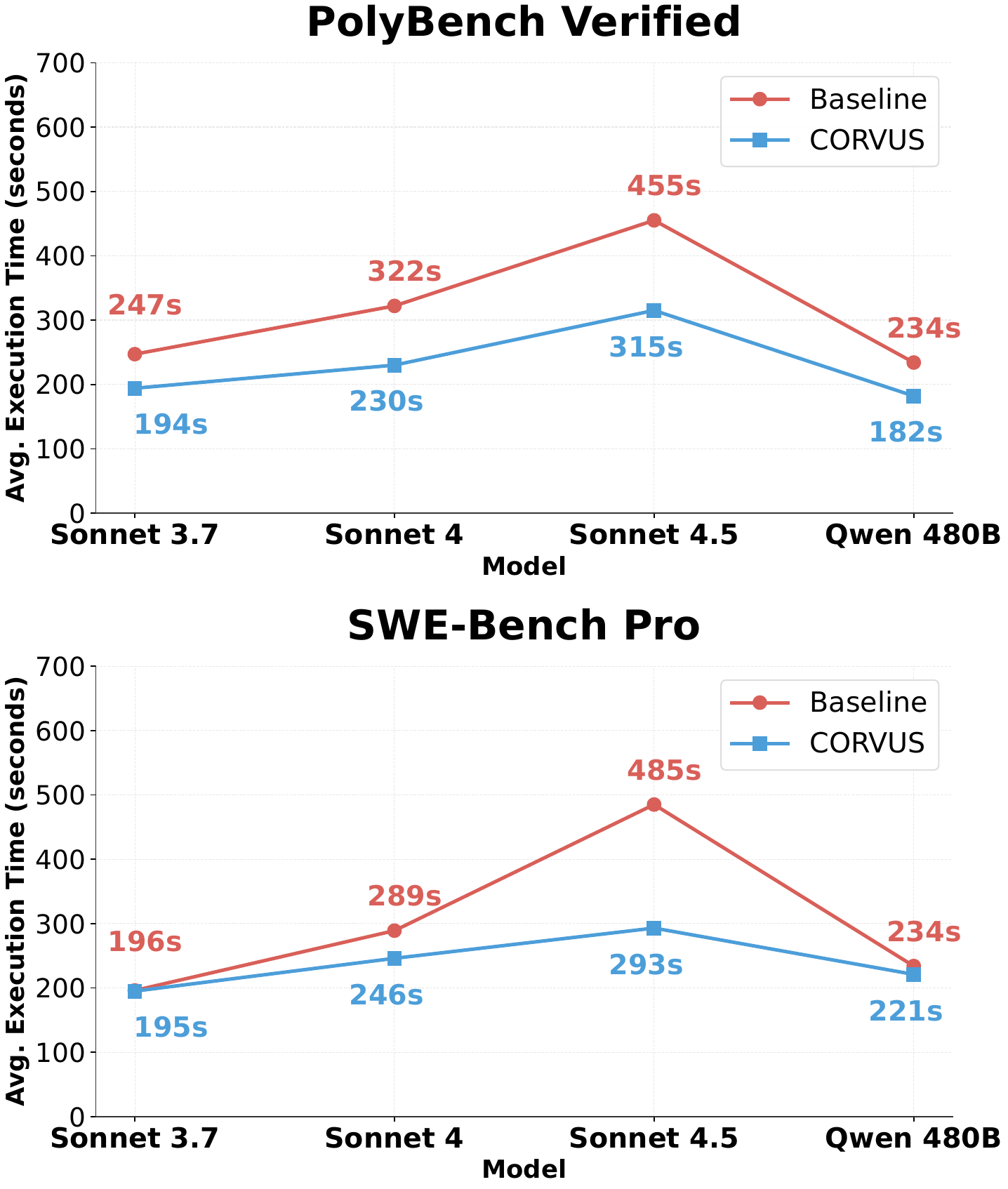}
        \vspace{-5mm}
        \caption{Execution time.}
        \vspace{-2mm}
        \label{fig:time}
    \end{minipage}
    
\end{figure*}

\begin{figure*}[!htb]
\includegraphics[width=\linewidth]{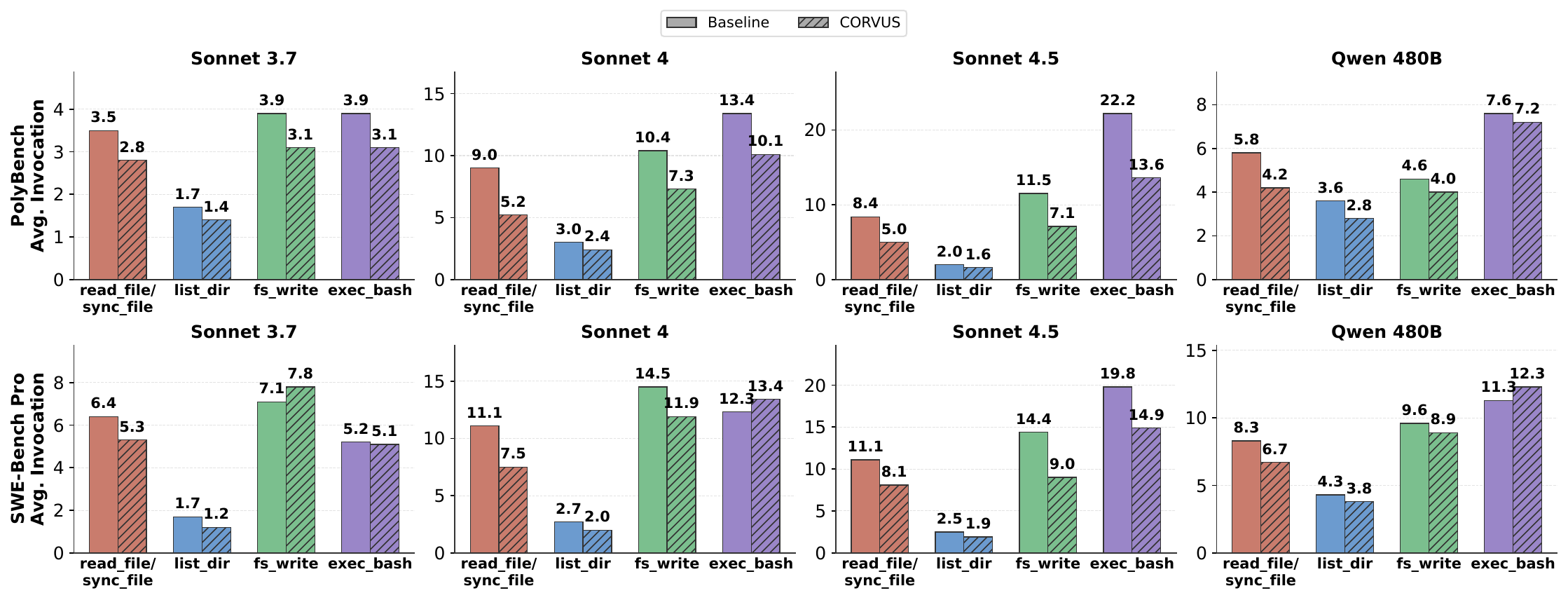}
\centering
\vspace{-5mm}
\caption{Average tool calls per instance.}
\label{fig:tool}
\end{figure*}

\textbf{Results and Analysis.} \sysname{} consistently improves trajectory efficiency across all models and benchmarks. We present the results in the order in which the effects arise: synchronized context first reduces repeated file access, then reduces the number of reasoning cycles, which together produce shorter final prompts and lower token cost.

\textbf{(1) Fewer Duplicate File Reads.}
As shown in \Cref{fig:reason}, \sysname{} reduces duplicate file reads by 22–86\% across all models. Since the \emph{synced context} is refreshed before each reasoning cycle, agents no longer need to repeatedly re-read files to recover updated contents, reducing redundant file access.

\textbf{(2) Fewer Reasoning Cycles.}
\Cref{tab:cycle} shows that \sysname{} completes tasks with fewer reasoning cycles for every evaluated model. The largest reduction is 37.33\% on \textsc{Claude Sonnet 4.5} for \textsc{SWE-PolyBench\_Verified}, decreasing from 45.03 to 28.22 cycles. To understand this reduction, we further measure tool invocations per instance.
As shown in \Cref{fig:tool}, \sysname{} substantially reduces \texttt{read\_file} operations. Other tool actions also decrease in most cases, suggesting that synchronized context reduces unnecessary exploration and recovery effort.

\begin{table*}[!htb]
\centering
\small
\caption{Reduction in reasoning cycles across two benchmarks.}
\setlength{\tabcolsep}{4pt}
\label{tab:cycle}
\begin{tabular}{lccc ccc}
\toprule
\multirow{2}{*}{\textbf{Model}} &
\multicolumn{3}{c}{\textsc{SWE-PolyBench\_Verified}} &
\multicolumn{3}{c}{\textsc{SWE-Bench Pro}} \\
\cmidrule(lr){2-4}\cmidrule(lr){5-7}
&
\textbf{Baseline} &
\textbf{\sysname{}} &
\textbf{Reduction} &
\textbf{Baseline} &
\textbf{\sysname{}} &
\textbf{Reduction} \\
\midrule
\textsc{Sonnet 3.7}
& 13.87 & 11.41 & \decrease{17.74\%}
& 21.43 & 20.38 & \decrease{4.90\%} \\

\textsc{Sonnet 4}
& 36.91 & 25.96 & \decrease{29.67\%}
& 41.69 & 35.81 & \decrease{14.10\%} \\

\textsc{Sonnet 4.5}
& 45.03 & 28.22 & \decrease{37.33\%}
& 48.64 & 34.81 & \decrease{28.43\%} \\

\textsc{Qwen 480B}
& 22.60 & 19.16 & \decrease{15.22\%}
& 34.51 & 32.71 & \decrease{5.22\%} \\
\bottomrule
\end{tabular}
\end{table*}

\textbf{(3) Shorter Final Trajectories.}
As shown in \Cref{fig:finalrequest}, \sysname{} shortens final request prompts by 15–32\% across models. This reduction comes from two sources.
First, \sysname{} avoids duplicate file snapshots: the baseline appends a new file snapshot to the trajectory on each repeated read, and retains older versions in history, whereas \sysname{} stores file contents in the \emph{synced context} with at most one current copy per synced file.
Second, fewer reasoning cycles reduce the number of accumulated tool calls and observations in message history, further reducing the final prompt length.

\textbf{(4) Lower Token Cost.}
The shorter trajectories also reduce total token usage and cost. As shown in \Cref{tab:token_cost}, \sysname{} reduces both input and output tokens across all models and benchmarks, with cost reductions up to 50.28\% on \textsc{Claude Sonnet 4.5} for \textsc{SWE-PolyBench\_Verified} (\$5.27 to \$2.62 per instance). These savings are the cumulative result of fewer duplicate file reads, fewer reasoning cycles, and shorter prompts throughout the trajectory.

\begin{table*}[!ht]
  \centering
  \caption{Token cost comparison between Baseline and \sysname{} on two benchmarks.}
 \small
  \setlength{\tabcolsep}{2pt}
 
    \begin{tabular}{l|rrr|rrr}
      \toprule
      \multirow{2}{*}{\textbf{Model}} &
      \multicolumn{3}{c|}{\textbf{Baseline}} &
      \multicolumn{3}{c}{\textbf{\sysname{}}} \\
      \cmidrule(lr){2-4} \cmidrule(lr){5-7}
       & 
      \textbf{Input Tokens} & \textbf{Output Tokens} & \textbf{Cost} &
      \textbf{Input Tokens} & \textbf{Output Tokens} & \textbf{Cost} \\
      \midrule
      \multicolumn{7}{c}{\textsc{SWE-PolyBench\_Verified}} \\
      \midrule
      \textsc{Sonnet 3.7} & 194,849 & 3,685.24 & \$0.64 & 133,721.54 \decrease{31.37\%} & 2,983.34 \decrease{19.05\%} & \$0.45 \decrease{29.69\%} \\
      \textsc{Sonnet 4} & 1,353,314.55 & 11,147.36 & \$4.23 & 746,981.43 \decrease{44.80\%} & 7,830.02 \decrease{29.76\%} & \$2.36 \decrease{44.21\%} \\
      \textsc{Sonnet 4.5} & 1,673,331.89 & 16,704.38 & \$5.27 & 829,579.15 \decrease{50.42\%} & 9,025.16~\decrease{45.97\%} & \$2.62 \decrease{50.28\%} \\
      \textsc{Qwen 480B} & 340,976.79 & 4,543.92 & \$0.08 & 246,139.48 \decrease{27.81\%} & 4,034.52 \decrease{11.21\%} & \$0.06 \decrease{25.00\%}  \\
      \midrule
      \multicolumn{7}{c}{\textsc{SWE-Bench Pro}} \\
      \midrule
      \textsc{Sonnet 3.7} & 406,413.33 & 7,586.92 & \$1.33 & 368,729.93 \decrease{\ \ 9.27\%} & 7,387.80 \decrease{\ \ 2.62\%} & \$1.22 \decrease{\ \ 8.27\%} \\
      \textsc{Sonnet 4} & 1,714,356.91 & 13,606.54 & \$5.35 & 1,397,655.99 \decrease{18.47\%} & 11,320.67 \decrease{16.80\%} & \$4.36 \decrease{18.50\%} \\
      \textsc{Sonnet 4.5} & 1,979,653.03 & 19,499.68 & \$6.23 & 1,119,821.14 \decrease{43.43\%} & 10,989.96 \decrease{43.64\%} & \$3.52 \decrease{43.50\%} \\
      \textsc{Qwen 480B} & 737,007.27 & 8,138.06 & \$0.18 & 627,603.96 \decrease{14.84\%} & 8,006.32 \decrease{\ \ 1.62\%} & \$0.15 \decrease{16.67\%} \\
      \bottomrule
    \end{tabular}%
  
  \label{tab:token_cost}
\end{table*}

\subsection{Runtime Efficiency and Synchronization Overhead}
\textbf{Setup and Metrics.}
We evaluate whether the additional context-sync step slows down execution. We report the average end-to-end runtime per instance as the primary metric. To explain synchronization overhead, we also measure the average number of files refreshed during each sync step, which approximates the size of the synchronization workload.

\begin{table*}[!htb]
\centering
\small
\setlength{\tabcolsep}{1.6pt}
\renewcommand{\arraystretch}{1.05}

\begin{minipage}[t]{0.6\textwidth}
\centering
\caption{Average synced file set size.}
\label{tab:synced_file_set}
\begin{tabular}{@{}lcc@{}}
\toprule
\textbf{Model} &
\textbf{\textsc{SWE-PolyBench\_Verified}} &
\textbf{\textsc{SWE-Bench Pro}} \\
\midrule
\textsc{Sonnet 3.7} & 2.66 & 5.01 \\
\textsc{Sonnet 4} & 4.84 & 7.14 \\
\textsc{Sonnet 4.5} & 4.58 & 7.81 \\
\textsc{Qwen 480B} & 3.64 & 6.35 \\
\bottomrule
\end{tabular}
\end{minipage}%
\hfill
\begin{minipage}[t]{0.38\textwidth}
\centering
\caption{Overall pass rate comparison.}
\label{tab:passrate}
\begin{tabular}{@{}lcc@{}}
\toprule
\textbf{Model} & \textbf{Baseline} & \textbf{\sysname{}} \\
\midrule
\textsc{Sonnet 3.7} & 19.20\% & \textbf{20.07\%} \passup{0.87} \\
\textsc{Sonnet 4} & \textbf{25.83\%} & 24.61\% \passdown{1.22} \\
\textsc{Sonnet 4.5} & 28.10\% & \textbf{29.49\%} \passup{1.39} \\
\textsc{Qwen 480B} & 20.07\% & \textbf{21.12\%} \passup{1.05} \\
\bottomrule
\end{tabular}
\end{minipage}

\end{table*}

\textbf{Results and Analysis.}
As shown in \Cref{fig:time}, \sysname{} reduces end-to-end runtime across all models and benchmarks, with up to 40\% reduction on \textsc{Claude Sonnet 4.5}.
Thus, the added synchronization step does not introduce observable runtime overhead. Instead, the savings from fewer reasoning cycles and shorter prompts dominate the cost of synchronization. This is expected because synchronization only performs local file reads to refresh registered files, which are typically milliseconds-level operations and much cheaper than LLM inference.

The synchronization workload remains small. As shown in \Cref{tab:synced_file_set}, each sync step refreshes only 2.66 to 4.84 files on average for \textsc{SWE-PolyBench\_Verified} and 5.01 to 7.81 files for \textsc{SWE-Bench Pro}.
This suggests that agents typically operate over a compact set of currently relevant files, so refreshing synced context adds little cost compared with LLM inference. Overall, \sysname{} improves runtime efficiency rather than trading token savings for synchronization overhead.

\subsection{Preserving Task Success}
\textbf{Setup and Metrics.}
Since \sysname{} is an efficiency optimization, we use pass@1 (the fraction of instances solved correctly on the first attempt without retries) as a guardrail metric to ensure that trajectory reductions do not substantially degrade task success.

\textbf{Results and Analysis.}
As shown in \Cref{tab:passrate}, \sysname{} achieves comparable pass@1 rates across all models. The key finding is that lighter-weight trajectories do not come at the cost of major task correctness: pass rates remain comparable, with three of four models showing slight improvements.

\subsection{Compatibility with Reactive Context Management}\label{subsec:compatibility}
\textbf{Setup and Metrics.}
Reactive methods reduce context after it has accumulated, while \sysname{} prevents duplicate and stale file snapshots from entering the trajectory. To test whether these mechanisms are complementary, we apply four reactive strategies (Sliding Window, Summarization, Observation Masking, and AgentDiet~\cite{DBLP:journals/corr/abs-2509-23586}) to both the baseline append-only agent and the \sysname{} synchronized agent.
We run all experiments on the 382 \textsc{SWE-PolyBench\_Verified} tasks using \textsc{Claude Sonnet 4.5}, and compare each \textsc{Baseline}+\textsc{Reactive} configuration with its corresponding \sysname{}+\textsc{Reactive} configuration.

\begin{table*}[!h]
\centering
\small
\caption{Compatibility between \sysname{} and reactive context-management methods. Deltas compare \sysname{}+\textsc{Reactive} against the corresponding \textsc{Baseline}+\textsc{Reactive}.
}
\label{tab:reactive_efficiency}
\setlength{\tabcolsep}{1.8pt}
\begin{tabular}{lcccccccc}
\toprule
\multirow{2}{*}{\textbf{Reactive Strat.}} &
\multicolumn{4}{c}{\textbf{Baseline + Reactive}} &
\multicolumn{4}{c}{\textbf{\sysname{} + Reactive}} \\
\cmidrule(lr){2-5}\cmidrule(lr){6-9}
&
\textbf{Cycles} &
\textbf{Input Tokens} &
\textbf{Cost} &
\textbf{Pass@1} &
\textbf{Cycles} &
\textbf{Input Tokens} &
\textbf{Cost} &
\textbf{Pass@1} \\
\midrule
None
& 41.5 & 1,453,143 & \$4.57 & 39.89\%
& 27.9 \decrease{32.8\%} & 796,236 \decrease{45.2\%} & \$2.53 \decrease{44.6\%} & 40.05\% \passup{0.16}\\

Sliding Window
& 40.3 & 1,206,749 & \$3.79 & 37.96\%
& 32.4 \decrease{19.6\%} & 902,115 \decrease{25.3\%} & \$2.85 \decrease{24.8\%} & 37.17\% \passdown{0.79}\\

Summarization
& 53.5 & 1,035,660 & \$3.51 & 39.53\%
& 29.6 \decrease{44.7\%} & 644,408 \decrease{37.8\%} & \$2.11 \decrease{39.9\%} & 36.65\% \passdown{2.88}\\

Obs. Masking
& 45.8 & 1,118,126 & \$3.57 & 34.55\%
& 27.2 \decrease{40.6\%} & 652,785 \decrease{41.6\%} & \$2.08 \decrease{41.7\%} & 36.39\% \passup{1.84}\\

AgentDiet
& 49.3 & 1,376,700 & \$4.37 & 32.72\%
& 27.5 \decrease{44.2\%} & 684,976 \decrease{50.2\%} & \$2.19 \decrease{49.9\%} & 36.65\% \passup{3.93}\\
\bottomrule
\end{tabular}
\end{table*}

\textbf{Results and Analysis.}
\Cref{tab:reactive_efficiency} shows that \sysname{} remains effective when combined with reactive context-management methods.
Across all reactive strategies, \sysname{}+\textsc{Reactive} uses fewer reasoning cycles (19.6\%-44.7\%), fewer input tokens (25.3\%-50.2\%), and lower cost (24.8\%-49.9\%) than the corresponding \textsc{Baseline}+\textsc{Reactive} configuration.
Pass-rate changes range from $-2.88$ to $+3.93$ percentage points, indicating broadly comparable task success.

Moreover, most combined configurations (\sysname{}+\textsc{Reactive}) are more efficient than \sysname{} alone. While \sysname{} reduces average input tokens from 1,453K (baseline) to 796K, combining it with Summarization, Observation Masking, or AgentDiet further reduces input tokens to 644K–685K. This suggests that the two types of methods are best used together: \sysname{} removes redundant or stale context upstream, while reactive methods compress residual trajectory content downstream. Overall, \sysname{} complements reactive context management rather than replacing it.

\section{Discussion}

\textbf{Limitations and Future Work.}\label{subsec:futurework}
Several extensions could make synchronized trajectories more adaptive. \textit{First}, a \texttt{desync\_file} tool would allow agents to remove files from the synced context once they become irrelevant, proactively shrinking the working set. \textit{Second}, partial synchronization could refresh selected functions or code blocks rather than entire files, though this requires robust region tracking across edits, e.g., AST-based anchoring~\cite{kim2026codestruct, tree-sitter, DBLP:conf/sp/ZhengXZ25, RFCAudit, DBLP:conf/kbse/FalleriMBMM14} or diff-aware boundary updates~\cite{gitdiff}. 
\textit{Third}, prompt caching can be combined with \sysname{} since stable prompt components like system instructions and tool schemas, remain cacheable.
However, synced context may change across cycles, and long-running operations such as tests or builds may exceed cache TTLs, limiting cache reuse for dynamic prompt components~\cite{promptcache}.
\textit{Finally}, fine-tuning models on \sysname{}-style trajectories could reduce the distribution gap between conventional append-only trajectories and decoupled synchronized context. Alignment methods such as on-policy distillation~\cite{song2026survey, DBLP:conf/iclr/AgarwalVZSGGB24, DBLP:journals/corr/abs-2601-18734} or reinforcement learning~\cite{DBLP:journals/tmlr/ZhangGYYZTZLXLZCZFWHVLW26, DBLP:journals/corr/abs-2412-05265} over decoupled trajectories may help models better exploit synchronized context, e.g., by maintaining smaller working sets.

\textbf{Human-AI Pair Programming.}
\sysname{} may reduce stale-context issues when developers and agents co-edit a repository. In append-only trajectories, once an agent reads a file, its snapshot remains in history. If a developer later modifies the same file, the agent may continue reasoning from the earlier snapshot and make edits against outdated code. With \sysname{}, synced files are refreshed at the start of each reasoning cycle, so developer edits will be reflected in the next cycle’s synced context. This helps the agent observe the latest repository state before making its next decision.

\section{Conclusion}
We introduced \sysname{}, an efficiency-oriented trajectory architecture for LLM coding agents. \sysname{} decouples file-read actions from file contents: instead of storing fixed snapshots in append-only trajectories, it maintains a synced file set and refreshes current contents before each reasoning cycle, preventing duplicate and stale snapshots by construction.
Our evaluation shows that \sysname{} produces lighter trajectories while preserving comparable pass rates, demonstrating that proactive synchronization improves the efficiency of long-horizon coding agents.



\bibliographystyle{unsrt}
\bibliography{reference}








\appendix
\newpage

\section{Experiments Compute Resources}\label{subsec:compute}
All experiments were conducted on a 16-inch MacBook Pro with an Apple M2 Pro processor and 16 GB of unified memory. LLM inference was performed through AWS Bedrock APIs; local compute was used only for running the agent framework, executing benchmark harnesses, managing repositories, and collecting logs. No local GPU training or fine-tuning was performed.

We accessed the evaluated models through AWS Bedrock\footnote{\url{https://aws.amazon.com/bedrock}} using the following model IDs:

\begin{itemize}

     \item[-] \textsc{Claude Sonnet 3.7}~\cite{anthropic_claude_3_7_sonnet}: \texttt{us.anthropic.claude-3-7-sonnet-20250219-v1:0}
        \item[-] \textsc{Claude Sonnet 4}~\cite{anthropic_claude_4_sonnet}: \texttt{us.anthropic.claude-sonnet-4-20250514-v1:0}
        \item[-]\textsc{Claude Sonnet 4.5}~\cite{anthropic_claude_4.5_sonnet}: \texttt{us.anthropic.claude-sonnet-4-5-20250929-v1:0}
        \item[-] \textsc{Qwen3-Coder-480B-A35B}~\cite{qwen3_coder}: \texttt{qwen.qwen3-coder-480b-a35b-v1:0}

\end{itemize}

\section{Task-Type Analysis}
\label{app:task_type_analysis}

We further analyze whether the efficiency gains of \sysname{} depend on the type of software-engineering task. We group tasks into three categories: bug fixing, feature implementation, and refactoring. We report results on both SWE-PolyBench Verified and SWE-Bench Pro across all evaluated models.

\begin{table*}[!ht]
\centering
\small
\caption{Task-type analysis on \textsc{SWE-PolyBench\_Verified} and \textsc{SWE-Bench Pro}.}
\label{tab:task_type_analysis}
\setlength{\tabcolsep}{3pt}
\renewcommand{\arraystretch}{1.12}
\begin{tabular}{llrrcc}
\toprule
\textbf{Model} &
\textbf{Category} &
\textbf{Baseline Tokens} &
\textbf{\sysname{} Tokens} &
\textbf{Token Reduction} &
\textbf{Cycle Reduction} \\
\midrule

\multicolumn{6}{c}{\textit{\textsc{SWE-PolyBench\_Verified}}} \\
\midrule

\multirow{3}{*}{\textsc{Sonnet 3.7}}
& Bug Fix     & 162,121   & 110,944   & \decrease{31.6\%} & \decrease{18.9\%} \\
& Feature     & 287,568   & 204,324   & \decrease{28.9\%} & \decrease{13.7\%} \\
& Refactoring & 448,343   & 277,433   & \decrease{38.1\%} & \decrease{21.1\%} \\
\addlinespace[3pt]

\multirow{3}{*}{\textsc{Sonnet 4}}
& Bug Fix     & 1,309,734 & 712,194   & \decrease{45.6\%} & \decrease{31.1\%} \\
& Feature     & 1,247,514 & 696,989   & \decrease{44.1\%} & \decrease{27.0\%} \\
& Refactoring & 2,925,369 & 1,813,607 & \decrease{38.0\%} & \decrease{17.8\%} \\
\addlinespace[3pt]

\multirow{3}{*}{\textsc{Sonnet 4.5}}
& Bug Fix     & 1,551,425 & 783,324   & \decrease{49.5\%} & \decrease{37.7\%} \\
& Feature     & 1,903,792 & 899,508   & \decrease{52.8\%} & \decrease{35.0\%} \\
& Refactoring & 3,236,245 & 1,516,914 & \decrease{53.1\%} & \decrease{41.2\%} \\
\addlinespace[3pt]

\multirow{3}{*}{\textsc{Qwen 480B}}
& Bug Fix     & 293,721   & 208,207   & \decrease{29.1\%} & \decrease{16.6\%} \\
& Feature     & 453,542   & 341,462   & \decrease{24.7\%} & \decrease{13.4\%} \\
& Refactoring & 821,732   & 605,301   & \decrease{26.3\%} & \increase{0.6\%} \\

\midrule
\multicolumn{6}{c}{\textit{\textsc{SWE-Bench Pro}}} \\
\midrule

\multirow{3}{*}{\textsc{Sonnet 3.7}}
& Bug Fix     & 375,970   & 304,572   & \decrease{19.0\%} & \decrease{7.6\%} \\
& Feature     & 420,632   & 424,447   & \increase{0.9\%}  & \decrease{3.6\%} \\
& Refactoring & 408,540   & 326,254   & \decrease{20.1\%} & \decrease{5.1\%} \\
\addlinespace[3pt]

\multirow{3}{*}{\textsc{Sonnet 4}}
& Bug Fix     & 1,683,789 & 1,255,281 & \decrease{25.4\%} & \decrease{17.0\%} \\
& Feature     & 1,637,317 & 1,231,815 & \decrease{24.8\%} & \decrease{15.6\%} \\
& Refactoring & 1,883,013 & 1,828,029 & \decrease{2.9\%}  & \decrease{9.2\%} \\
\addlinespace[3pt]

\multirow{3}{*}{\textsc{Sonnet 4.5}}
& Bug Fix     & 1,989,294 & 1,086,304 & \decrease{45.4\%} & \decrease{29.2\%} \\
& Feature     & 1,958,052 & 1,148,888 & \decrease{41.3\%} & \decrease{26.2\%} \\
& Refactoring & 2,010,159 & 1,097,705 & \decrease{45.4\%} & \decrease{31.6\%} \\
\addlinespace[3pt]

\multirow{3}{*}{\textsc{Qwen 480B}}
& Bug Fix     & 816,978   & 684,848   & \decrease{16.2\%} & \decrease{0.2\%} \\
& Feature     & 706,872   & 651,976   & \decrease{7.8\%}  & \decrease{8.4\%} \\
& Refactoring & 718,262   & 722,458   & \increase{0.6\%}  & \increase{0.2\%} \\

\bottomrule
\end{tabular}
\end{table*}

The results show that \sysname{} provides consistent efficiency gains across task types and benchmarks.
On \textsc{SWE-PolyBench Verified}, \sysname{} reduces input tokens by 24.7–53.1\% across all model/category pairs, with similarly consistent cycle reductions; the only exception is a negligible 0.6\% cycle increase for \textsc{Qwen3-Coder-480B-A35B} on refactoring tasks. On \textsc{SWE-Bench Pro}, \sysname{} also reduces tokens and cycles in most settings, with especially large gains for \textsc{Claude Sonnet 4.5}. The few regressions are small, such as a 0.9\% token increase for \textsc{Claude Sonnet 3.7} on feature tasks and a 0.6\% token increase for \textsc{Qwen3-Coder-480B-A35B} on refactoring tasks. Overall, the gains are not driven by a single task category, suggesting that \sysname{} is broadly effective across different types of coding tasks.

\section{Ablation on Synced Context Placement}
\label{app:placement_ablation}

We study where the synced context should be placed in the prompt. In \sysname{}'s default design, synced context is inserted after the message history, so the refreshed repository state appears immediately before the next model invocation. We compare this with an alternative placement that inserts the same synced context immediately after the system prompt, before the message history. Both variants are evaluated on the same 50 instances using \textsc{Claude Sonnet 4.5}.

\begin{table}[h]
\centering
\small
\caption{Ablation on synced context placement. \textsc{After-System-Prompt} inserts synced context after the system prompt but before the message history, while \textsc{After-Message-History} is the default placement used in \sysname{}, inserting synced context right after the message history.}
\label{tab:placement_ablation}
\vspace{3mm}
\setlength{\tabcolsep}{3.5pt}
\renewcommand{\arraystretch}{1.05}

\begin{tabular}{l r@{\;}l r@{\;}l r@{\;}l}
\toprule
\textbf{Metric} &
\multicolumn{2}{c}{\textbf{Baseline}} &
\multicolumn{2}{c}{\textbf{\textsc{After-System-Prompt}}} &
\multicolumn{2}{c}{\textbf{\textsc{After-Message-History}}} \\
\midrule

Pass Rate
& 31.71 & \nodiff
& 34.15 & \passup{2.44}
& \textbf{34.88} & \passup{3.17} \\

Avg. Cycles
& 45.7 & \nodiff
& 35.5 & \decrease{22.4\%}
& \textbf{29.8} & \decrease{34.9\%} \\

Avg. Tokens
& 1,609,534 & \nodiff
& 1,023,841 & \decrease{36.4\%}
& \textbf{872,595} & \decrease{45.8\%} \\

Avg. Cost
& \$5.06 & \nodiff
& \$3.22 & \decrease{36.4\%}
& \textbf{\$2.75} & \decrease{45.6\%} \\

\bottomrule
\end{tabular}
\end{table}

Both synced-context placements improve over the append-only baseline, confirming that decoupling file contents from chronological history is beneficial regardless of placement. However, placing synced context after the message history is consistently more efficient than placing it near the system prompt: it reduces cycles by 34.9\% rather than 22.4\%, tokens by 45.8\% rather than 36.4\%, and cost by 45.6\% rather than 36.4\%. Pass rates remain comparable across the two placements. These results support our default design: refreshed repository state is most useful when placed closest to the model's next decision, after the accumulated history it is meant to update.

\section{Broader Impact}
\label{sec:broader_impact}

\sysname{} is an efficiency-oriented architecture for LLM coding agents. Its primary intended impact is to reduce the cost of long-horizon software-engineering workflows by avoiding redundant file snapshots and keeping file context synchronized with the current repository state. This can make coding agents more practical for tasks that require many reasoning cycles, especially when inference cost or context-window limits are a bottleneck. Reducing unnecessary input tokens and recovery cycles may also reduce the computational cost of running agentic coding systems.

The main societal risks are those of coding agents more broadly. Making agentic workflows cheaper and faster can support legitimate software development, but it may also lower the cost of producing buggy, insecure, or harmful code. \sysname{} does not introduce new code-generation capabilities, vulnerability-discovery methods, or autonomous deployment mechanisms; it changes how repository context is maintained. Nevertheless, efficiency improvements can amplify both beneficial and harmful uses of existing coding agents.




\newpage

\end{document}